\documentclass{article}
\usepackage{spconf,amsmath,graphicx,hyperref}
\usepackage{cite}
\usepackage{amssymb,amsfonts}
\usepackage{algorithmic}
\usepackage{textcomp}
\usepackage{xcolor}
\usepackage{booktabs}
\usepackage{subcaption}

\title{Attention-Based Adaptive Policies for Simultaneous Speech-to-Text Translation}
\name{Filip Tășădan, Ema Tomanová, Ondrej Lopuch, Paweł Bilko, Anders Søgaard}
\address{University of Copenhagen \\ tasadan.filip@gmail.com}
\begin{document}
%
\maketitle
%


\begin{abstract}
\par Simultaneous speech-to-text translation (Simul-S2TT) consists of generating partial translations while the incoming audio frames are processed by the system. However, the streaming nature of this setup creates the challenge of deciding the best moment to perform an accurate translation while minimizing the delay. To address this challenge, we utilize the cross-attention mechanism of the encoder-decoder architecture to find the right alignment between the input speech frames and the target text tokens. In this paper, we propose the \textit{Recent Frame Attention Policy (RFAP)} and the \textit{Dual-Condition Attention Policy (DCAP)} that allow offline trained speech-to-text translation models to be used in streaming scenarios without requiring additional training. Results on three different language translation pairs over the CVSS-C corpus show that the RFAP is able to surpass other policies with gains of up to 4.0 BLEU while reducing the translation delay by almost 1 second. Moreover, the DCAP is able to preserve a high translation quality when the latency is very low.
\end{abstract}

\section{Introduction}
\par Simultaneous speech-to-text translation (Simul-S2TT) is used for real-time translation of international conferences and live streams in various languages \cite{fugen2007simultaneous} \cite{bangalore2012real}. In order to balance the translation quality with latency, a decision policy is employed that guides the system when to \textbf{READ} (wait for more audio) or \textbf{WRITE} (output a partial translation for the accumulated input chunks) \cite{ma2020simuleval}. 
\par The simultaneous decision policies are categorized into two main types: \textit{fixed} and \textit{adaptive}. Fixed policies, such as the widely used Wait-$k$ approach, rely on simple heuristics where the model waits for a predetermined number of speech segments before generating the output \cite{nguyen2021empirical}. The main limitation of the fixed policy models is that they cannot determine whether at each time step they have enough information context to be able to generate an accurate translation or not, thus leading to suboptimal performance \cite{liu2024recent}. 
\par In order to mitigate this issue, adaptive decision policies have been proposed. These policies determine the moment of translation by finding correlations between the input context and the translation output \cite{sethiya2025end}. The goal of the adaptive policies is to provide the Simul-S2TT models with a mechanism that dynamically determines the moments when the accumulated information is sufficient for the model to accurately produce a translation. On the other hand, if the adaptive decision mechanism has any indication that the current input buffer does not contain enough information to reliably translate it (i.e. partial words, incomplete expressions), it will signal to the model that it must wait for more information to be received before translating it. The issue with most of the models that have built-in adaptive decision components is that they are provided with partial input during training to simulate the real-time translation conditions \cite{ma2020simulmt} \cite{zhang2022gaussian} \cite{ma2023efficient}, thus increasing the computation costs and making the models more prone to errors when training and testing conditions are different from each other.
\par To avoid simulating the streaming conditions during training, adaptive policies applied to offline-trained models have been proposed by \cite{papi2023attention}, \cite{papi2023alignatt} and \cite{yan2023cmu}. The main advantage of these policies is that they do not require any additional modules for the decision policy. In the case of \cite{papi2023attention} and \cite{papi2023alignatt} the policy is determined by the attention-heads alignment between the input speech and the generated output. In \cite{yan2023cmu}, the decision policy is computed by performing a beam search and selecting the best hypothesis for each chunk.
\par In this paper, we propose the Recent Frame Attention Policy (RFAP) and the Dual-Condition Attention Policy (DCAP) that allow standard offline speech-to-text models to be used in streaming scenarios. Our research can be summarized as follows:
\begin{itemize}
    \item We show that RFAP outperforms all the other evaluated policies by gaining up to 4.0 BLEU while reducing the delay by almost 1 second.
    \item We show that the DCAP is capable of achieving negative average latency with high translation quality
\end{itemize}

\section{Preliminary}
\par Our adaptive decision policies use the cross-attention mechanism of the decoder module to determine the right translation moment \cite{vaswani2017attention}. This mechanism enables the model to learn the relationship between the encoded speech tokens and the generated text tokens. The attention score is computed as:
\begin{equation}
\operatorname{Attention}(Q, K, V) = \operatorname{softmax}\!\left(\frac{QK^{\top}}{\sqrt{d_k}}\right) V
\label{eq:attention}
\end{equation}

\par To compute the cross attention for our models, at each time step $t$, the Simul-S2TT model receives $n$ speech frames $X_n=[x_1, x_2, ..., x_n]$ and it has previously generated the output sequence $Y_{m-1}=[y_1, y_2, ..., y_{m-1}]$. The input sequence $X_n$ is further used to compute the key ($K^{t}$) and value ($V^{t}$) matrices. For the first cross-attention layer, the query ($Q^{t}$) is obtained from the generated text sequence $Y_{m-1}$. For the subsequent attention layers, $Q^{t}$ is obtained from the output of the previous attention layer. Thus, the cross-attention at a given time step $t$ can be regarded as a function of $X_n$ and $Y_{m-1}$, defined as:
\begin{equation}
\lambda^{t}(X_n, Y_{m-1}) = \operatorname{softmax}\!\left(\frac{Q^{t}(K^{t})^{\top}}{\sqrt{d_k}}\right) V^{t}
\label{eq:attention}
\end{equation}

\par  As observed in \cite{papi2023attention}, when the attention score for the generated output is concentrated on the most recent speech frames, the model requires more context before emitting a new translation. On the other hand, when the attention of the generated output is focused on older frames, the model has accumulated enough information to generate the translation. In order to quantify the relationship between the source speech frames and the generated translation, we focus on the recent frames attention, defined as the sum of attention weights allocated to the $r$ most recently received speech frames, i.e. the frames of the latest input chunk, for the target token $y_m$:
\begin{equation}
\lambda_{\text{recent}}^{t}(X_n, y_m) = \sum_{i=n-r+1}^{n} \lambda^{t}(x_i, y_m)
\label{eq:recent_attention}
\end{equation}
\par Since $y_m$ is the text token generated at time step $t$, we have to employ a two-pass inference on the auto-regressive decoder in order to compute the cross attention score as presented in Equation \ref{eq:recent_attention}. First, the encoded speech frames $X_n$ together with the previously generated tokens $Y_{m-1}$ are provided to the decoder in order to compute the output token $y_m$. Afterwards, we run again the inference on the decoder having as input the same $X_n$ but the output text sequence becomes $Y_m$ since we append the token $y_{m}$ to $Y_{m-1}$. In this second pass, we are able to extract from the decoder the cross attention score as defined by Equation \ref{eq:recent_attention} and further use it for the decision policies presented by the next section.

\section{Method}
\subsection{Recent Frame Attention Policy (RFAP)}
\label{rfap-policy-section}
\par The \textbf{Recent Frame Attention Policy} is built on the hypothesis that the cross attention mechanism shifts its focus from the most recent audio frames once it has gathered enough information \cite{papi2023attention}. The RFAP compares the sum of the attention weights $\lambda_{\text{recent}}^{t}(X_n, y_m)$  of the most recently received frames for the generated text token $y_m$ against a fixed threshold $\alpha$:
\begin{equation}
\lambda_{\text{recent}}^{t}(X_n, y_m) < \alpha, \alpha \in (0,1)
\label{recent_attention}
\end{equation}

\begin{figure}[h]
    \centering
    \includegraphics[width=0.74\columnwidth]{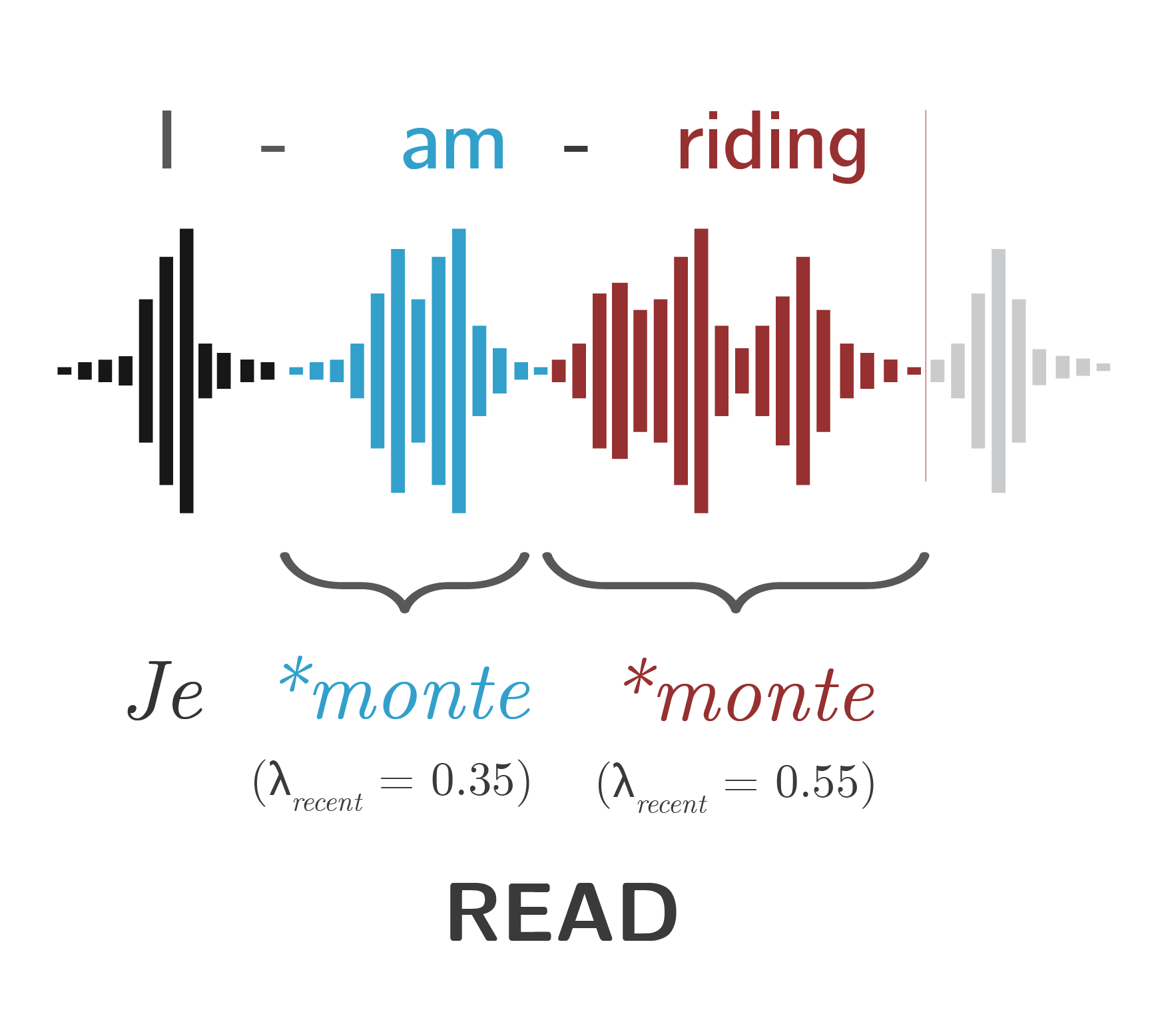}
    \caption{\textsf{READ Decision}: High recent attention score ($\lambda_{\text{recent}}^{t} = 0.55 > \alpha, \alpha=0.2$) indicates the model requires more source context. Recent frames are denoted in red, previous frames with lambda over threshold in blue and previous frames that invoked \textbf{WRITE} decision in black. Future source frames in light gray. Next target token prediction marked with *.}
    \label{fig:read_action}
\end{figure}

\begin{figure}[h]
    \centering
    \includegraphics[width=0.74\columnwidth]{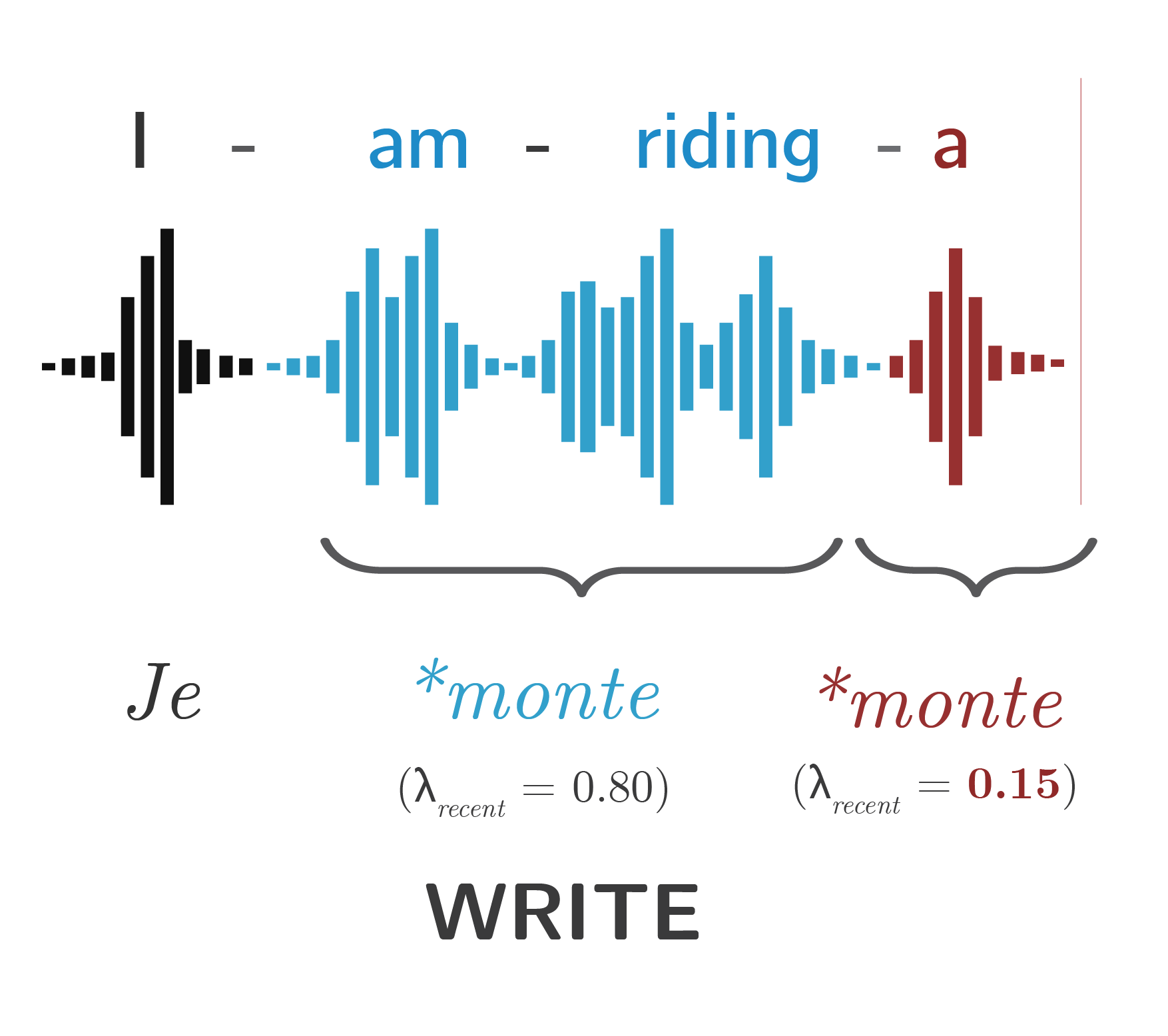}
    \caption{\textsf{WRITE Decision}: Focus shifts to earlier context ($\lambda_{\text{recent}}^{t} = 0.15 < \alpha, \alpha=0.2$), signaling sufficient information to emit the target text token 'monte'.}
    \label{fig:write_action}
\end{figure}

\par Following Eq. \ref{recent_attention},  when the $\lambda^{t}_{\text{recent}}$ score is higher than the threshold $\alpha$, the incoming audio chunks are strongly correlated with the generated output, so the system has to wait for more context before translating. In this case, the \textbf{READ} action will be emitted by the policy as depicted in Figure \ref{fig:read_action}. Otherwise, when the recent attention score is lower than the threshold $\alpha$, the necessary information to translate token $y_m$ has a lower correlation with the most recently received speech frames. Thus, the policy will trigger the \textbf{WRITE} action as in Figure \ref{fig:write_action}. The threshold $\alpha$ can be used to balance the translation quality and latency. 

\par Due to the two-pass inference strategy, the RFAP only computes the cross-attention score between the next output token and the most recent speech frames, thus preserving the same hypothesis as \cite{papi2023attention} while eliminating the need for a excessive accumulation of speech frames before emitting the WRITE action.

\subsection{Dual-Condition Attention Policy (DCAP)}
\par The RFAP waits for the attention score to stabilize before triggering the \textbf{WRITE} action, which improves the translation quality but increases the delay. In order to mitigate this issue, we propose the \textbf{Dual-Condition Attention Policy} that extends the RFAP by considering not only the correlation between the speech frames and the generated translation tokens, but also the rate of change in the attention focus on the most recent frames. The information provided by the rate of change allows the model to trigger the \textbf{WRITE} action whenever new tokens that impact the translation are being read by the model.

\par To calculate the rate of change, we look at the difference of the recent attention score at the current time step $t$ and the previous time step $t-1$ for the target token $y_m$. We define this relationship as:
\begin{equation}
    \Delta \lambda_{\text{recent}}(X_n, y_m) = \lambda_{\text{recent}}^{t}(X_n, y_m)  - \lambda_{\text{recent}}^{t-1}(X_{n-r}, y_{m}) 
   \label{eq:attention_rot}
\end{equation}

\par The DCAP policy is built on the idea that the cross-attention mechanism exhibits two phases where the translation can be made: the \textit{Discovery Phase} and the \textit{Completion Phase}. In order to capture both phases, the DCAP policy uses a dual-condition as presented in Equation \ref{eq:dual_condition}:
\begin{equation}
     \Delta \lambda_{\text{recent}}(X_n, y_m) \geq 0 \quad \lor \quad \lambda_{\text{recent}}^{t}(X_n, y_m) < \alpha
    \label{eq:dual_condition}
\end{equation}

\par The \textbf{rate of change condition} ($\Delta \lambda_{\text{recent}}(X_n, y_m) \geq 0$) acts as a trigger for the \textit{Discovery Phase}. A positive rate of change shows that the model is finding relevant information in the new audio frames. Instead of waiting for the attention score to peak and then drop, this condition lets the system produce the translation as soon as it detects the new information.

\par The \textbf{recent attention condition} ($\lambda_{\text{recent}}^{t}(X_n, y_m) < \alpha$) acts as a trigger for the \textit{Completion Phase}. This rule ensures that the translation is triggered once the decoder shifts its focus away from the most recent frames, particularly in cases when the rate of change condition is not triggered because the recent attention has a descending curve and it eventually gets below the threshold $\alpha$. In this case, the model has all the relevant context for translation.

\begin{figure*}[ht]
\centering
\begin{subfigure}[b]{0.32\textwidth}
    \centering
    \includegraphics[width=\linewidth]{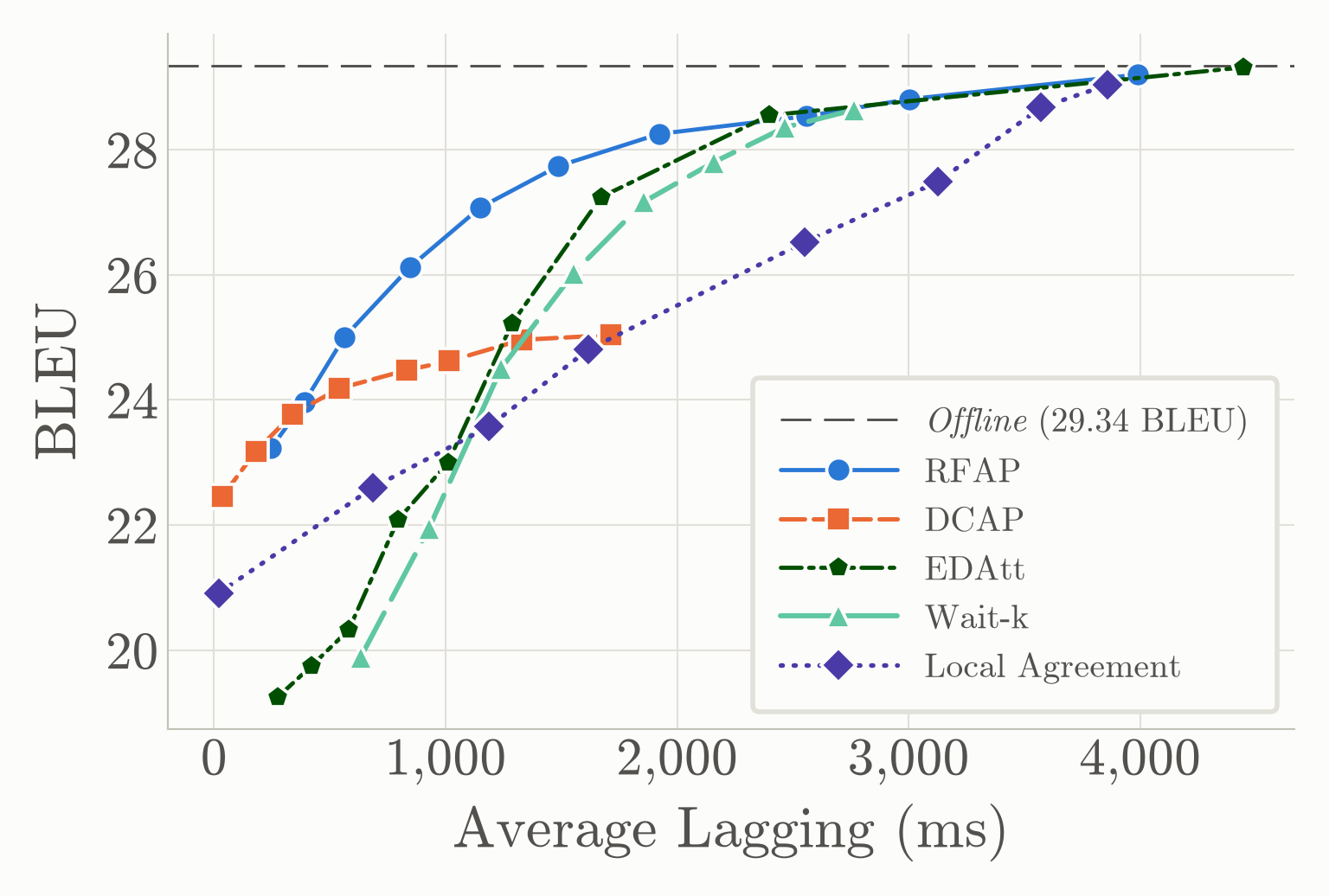}
    \caption{FR-EN}
    \label{fig:bleu_vs_al_fr}
\end{subfigure}\hfill
\begin{subfigure}[b]{0.32\textwidth}
    \centering
    \includegraphics[width=\linewidth]{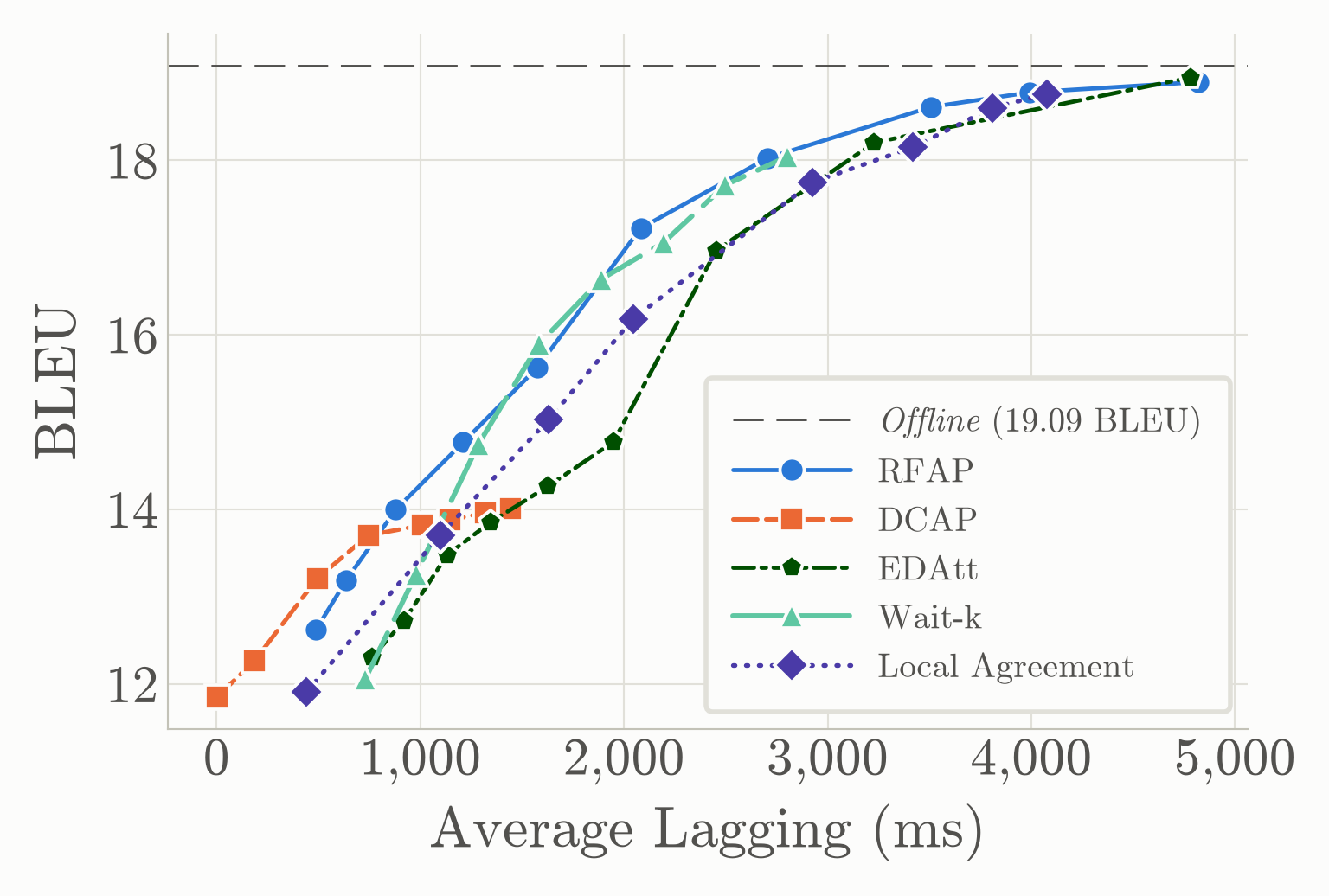}
    \caption{DE-EN}
    \label{fig:bleu_vs_al_de}
\end{subfigure}\hfill
\begin{subfigure}[b]{0.32\textwidth}
    \centering
    \includegraphics[width=\linewidth]{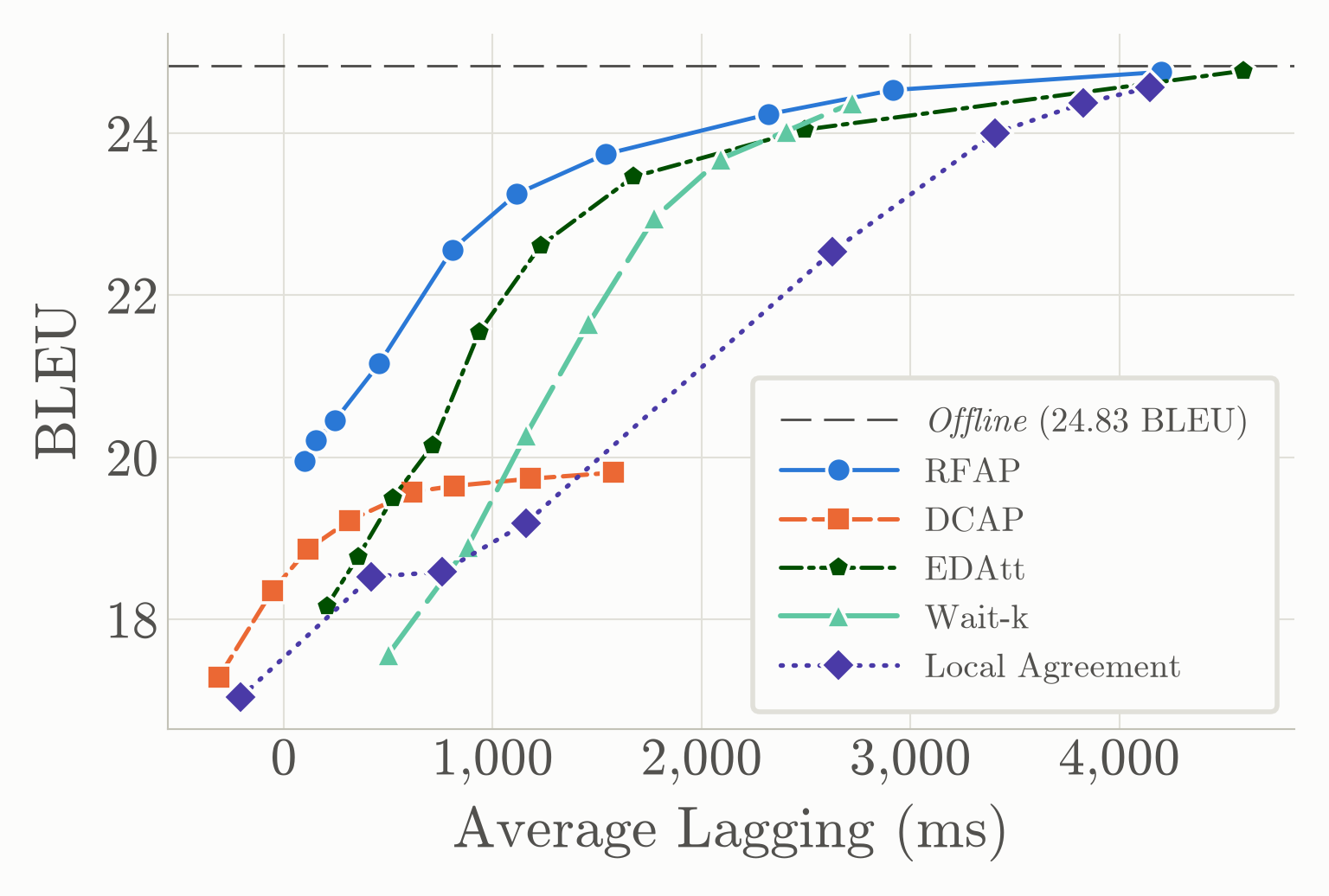}
    \caption{ES-EN}
    \label{fig:bleu_vs_al_es}
\end{subfigure}
\caption{BLEU score vs. Average Lagging (AL) for the (a) FR-EN, (b) DE-EN, and (c) ES-EN translation pairs.}
\label{fig:bleu_vs_al}
\end{figure*}

\section{Experiments}
\subsection{Experimental Setup}
\begin{table}[h]
\centering
\caption{CVSS-C Dataset Statistics}
\label{dataset_statistics}
\resizebox{\columnwidth}{!}{%
\begin{tabular}{llrrrrrr}
\toprule
\textbf{Language Pair} & \textbf{Split} & \textbf{No. Samples} & \textbf{Hours} & \textbf{Avg. SRC Duration (s)} & \textbf{Avg. TGT Tokens Count} \\
\midrule
FR-EN      & Train  & 207,364 & 174.0 &   4.57 &  11.6 \\
           & Dev    &  14,759 &  13.0 &   5.28 &  12.4 \\
           & Test   &  14,759  &  13.3 &   5.66 &  12.7 \\
\midrule
DE-EN      & Train  & 127,822 & 112.4 & 5.17 & 12.3 \\
           & Dev    & 13,511 & 12.5 & 5.48 & 13.0 \\
           & Test   & 13,504 & 12.1 & 5.72 & 12.7 \\
\midrule
ES-EN      & Train  & 79,012 & 69.5 & 5.13 & 12.0 \\
           & Dev    & 13,212 & 12.4 & 5.92 & 12.9 \\
           & Test   & 13,216 & 12.4 & 6.17 & 12.9 \\
\bottomrule
\end{tabular}%
}
\end{table}

\par For our experiments, we utilized the CVSS-C dataset \cite{jia2022cvss} on the French-English (FR-EN), German-English (DE-EN), Spanish-English (ES-EN) language pairs. This dataset is derived from the CoVoST 2 speech-to-text translation corpus \cite{wang2021covost}, that consists of synthetically generated input speech in a single canonical voice. Table~\ref{dataset_statistics} presents the distribution of the data in the CVSS-C dataset.

\par Our speech-to-text translation model is based on the StreamSpeech architecture \cite{zhang2024streamspeech} adapted for offline speech-to-text translation. The model has a size of 55 million parameters. The encoder follows the Conformer architecture \cite{gulati2020conformer}, consisting of 12 layers, 4 attention heads, an embedding dimension of 256, and a feed-forward dimension of 2048. The decoder is a standard Transformer \cite{vaswani2017attention} with 4 layers, 8 attention heads, an embedding dimension of 512, and a feed-forward dimension of 2048. The encoder produces one speech frame every 40 ms, and the input is received in fixed chunks of 320 ms. We therefore set the number of the most recent speech frames $r = 8$. Moreover, for the attention-based decision policies, we extract $\lambda_{\text{recent}}^{t}$ from the last cross-attention layer, which we empirically found to yield the best results.
\par The model was trained on an NVIDIA RTX 4000 GPU with 31GB of memory for 80 epochs. Training was conducted with mixed-precision (FP16) to improve computational efficiency.

\subsection{Results}
\label{results}
\par In this section, we present the experimental results of the \textbf{RFAP} and the \textbf{DCAP} policies on the FR-EN, DE-EN and ES-EN language translation pairs from the CVSS-C dataset \cite{jia2022cvss}. The main goal is to evaluate how well these attention-based policies balance the trade-off between translation quality and latency. The translation quality is measured by the BLEU score \cite{papineni2002bleu}, whereas the delay is computed as average latency (AL) measured in milliseconds \cite{ma2019stacl}. We compare our proposed decision policies against \textit{EDAtt} \cite{papi2023attention}, and two baseline policies, namely \textit{Wait-K} \cite{ma2019stacl} and \textit{Local Agreement} (LA) \cite{liu2020low}. In Fig. \ref{fig:bleu_vs_al}, we report the evaluation results of the proposed policies in terms of latency-quality trade-off. Table~\ref{tab:policy_bleu_by_al_bin} groups the operating points of each policy from Fig.~\ref{fig:bleu_vs_al} into AL ranges and reports the best BLEU reached within each range, allowing a direct comparison of the policies under the same latency constraints.

\begin{table}[t]
\centering
\caption{Best BLEU reached by each policy within a given AL range. \textendash{} marks a range where policy has no result.}
\label{tab:policy_bleu_by_al_bin}
\scriptsize
\setlength{\tabcolsep}{2.5pt}%
\begin{tabular*}{\columnwidth}{@{\extracolsep{\fill}}l*{8}{c}@{}}
\toprule
& \multicolumn{8}{c}{\textbf{AL range (s)}} \\
\cmidrule(l){2-9}
\textbf{Policy} & $<$0 & 0--0.5 & 0.5--1 & 1--1.5 & 1.5--2 & 2--2.5 & 2.5--3 & $>$3 \\
\midrule
\multicolumn{9}{l}{\textit{French-English (FR-EN)}} \\
RFAP & -- & \textbf{23.97} & \textbf{26.12} & \textbf{27.73} & \textbf{28.25} & -- & 28.53 & 29.20 \\
DCAP & -- & 23.77 & 24.48 & 24.96 & 25.04 & -- & -- & -- \\
EDAtt & -- & 19.75 & 22.08 & 25.22 & 27.24 & \textbf{28.55} & -- & \textbf{29.31} \\
Wait-K & -- & -- & 21.95 & 24.50 & 27.17 & 28.35 & \textbf{28.64} & -- \\
Local Agreement & -- & -- & 22.60 & 23.57 & 24.80 & -- & 26.52 & 29.03 \\
\midrule[\heavyrulewidth]
\multicolumn{9}{l}{\textit{German-English (DE-EN)}} \\
RFAP & -- & 12.62 & \textbf{14.00} & \textbf{14.77} & 15.62 & 17.22 & 18.02 & 18.89 \\
DCAP & -- & \textbf{13.21} & 13.70 & 14.01 & -- & -- & -- & -- \\
EDAtt & -- & -- & 12.72 & 13.85 & 14.77 & 16.97 & -- & \textbf{18.95} \\
Wait-K & -- & -- & 13.26 & 14.74 & \textbf{16.64} & \textbf{17.72} & \textbf{18.04} & -- \\
Local Agreement & -- & 11.90 & -- & 13.71 & 15.03 & 16.18 & 17.75 & 18.76 \\
\midrule[\heavyrulewidth]
\multicolumn{9}{l}{\textit{Spanish-English (ES-EN)}} \\
RFAP & -- & \textbf{21.16} & \textbf{22.56} & \textbf{23.25} & \textbf{23.74} & \textbf{24.23} & \textbf{24.53} & 24.75 \\
DCAP & \textbf{17.29} & 19.22 & 19.65 & 19.73 & 19.81 & -- & -- & -- \\
EDAtt & -- & 18.77 & 21.54 & 22.61 & 23.46 & 24.04 & -- & \textbf{24.77} \\
Wait-K & -- & 17.57 & 18.90 & 21.65 & 22.95 & 24.02 & 24.37 & -- \\
Local Agreement & 17.04 & 18.52 & 18.59 & 19.19 & -- & -- & 22.54 & 24.56 \\
\bottomrule
\end{tabular*}
\end{table}

\par Table \ref{tab:policy_bleu_by_al_bin} shows RFAP as strong contender across all latencies. In the mid-range latencies it performs better than any other policy except for wait-k in DE-EN dataset. When compared to EDAtt, the RFAP is capable to preserve high BLEU score even with latency lower by 500 ms on average. In high-latency settings RFAP keeps up on par with all other policies. On the FR-EN translation, the RFAP gains $\approx$4.0 BLEU points compared to EDAtt when the AL is below 500ms. This result experimentally proves what was stated in Section \ref{rfap-policy-section}, i.e. that RFAP eliminates the need for a large accumulation of speech frames before emitting the \textbf{WRITE} action, thus preserving a very strong translation quality even when the latency is low. The RFAP achieves a decrease in latency of around 1 second when compared to \textit{Wait-K} and 2 seconds to \textit{LA} on the FR-EN and ES-EN translations for similar translation qualities.
\par The nature of DCAP allows it to excell in low-latency settings. The rate of change mechanism employed by this policy makes it generate the output translation very soon, but it also leads to degraded performance as the latency is increased. On the FR-EN translation, the DCAP is able to achieve a BLEU score of 22.46 with only 36ms AL. Moreover, on the ES-EN translation, the DCAP is able to achieve negative latency (AL = -311ms) while preserving a translation quality of 17.29 BLEU. Since the speech input chunk size is fixed to 320ms, we can infer that on the ES-EN translation, DCAP is able on average to translate with almost one chunk ahead of the target reference.

\section{Conclusion}
\par In this paper, we proposed the Recent Frame Attention Policy and the Dual-Condition Attention Policy, two simultaneous translation decision policies that leverage the cross-attention mechanism of offline-trained speech-to-text models. Our experimental results show that the RFAP achieves a superior translation quality in mid-range latency settings compared to existing policies, outperforming the EDAtt, Wait-$k$ and Local Agreement policies with gains of up to 4.0 in BLEU score. On the other hand, the DCAP is highly effective in low-latency settings, since it is capable of achieving negative AL while preserving a strong translation quality. In conclusion, this research demonstrates that attention-based policies enable offline models to balance translation quality and latency in real-time streaming scenarios without requiring additional training or complex decision components.

\section{Compliance with Ethical Standards}
\par This research was conducted using the publicly available CVSS-C corpus \cite{jia2022cvss}, derived from CoVoST 2 \cite{wang2021covost}. No new data involving human subjects were collected, and ethical approval was not required.

\bibliographystyle{IEEEbib}
\bibliography{strings,refs}

\end{document}